\documentclass[conference]{IEEEtran}
\usepackage{cite}
\usepackage[pdftex]{graphicx}
\usepackage[cmex10]{amsmath}
\usepackage{amssymb}

\usepackage{times}
\usepackage{array}
\usepackage[tight,footnotesize]{subfigure}
\usepackage[font=footnotesize]{subfig}
\usepackage{graphicx}
\usepackage{mathtools}
\usepackage{microtype}
\usepackage{amsmath}

\usepackage{alltt}
\usepackage{shadethm}
\newshadetheorem{comment}{Comment}

\def\boxx{{\vcenter{\vbox{\hrule height.3pt
          \hbox{\vrule width.3pt height6pt
          \kern6pt\vrule width.3pt}\hrule height.3pt}}\;}}

\def\impos{{\;\vcenter{\hbox{\rule{5mm}{0.2mm}}} \vcenter{\hbox{\rule{1.5mm}{1.5mm}}} \;}}

\def\lrarrow{\leftrightarrow \kern-8pt \rightarrow}

\def\2{\frac{1}{2}}

\def\beq{\begin{eqnarray}}
\def\eeq{\end{eqnarray}}
\def\2{\frac{1}{2}}

\def\lrarrow{\leftrightarrow \kern-8pt \rightarrow}

\def\frightarrow{\rightarrow \kern-11pt /~~}
\def\reducesto{\simeq \kern -3pt >}

\usepackage{fancyhdr}					
\begin{document}
\newcommand{\strust}[1]{\stackrel{\tau:#1}{\longrightarrow}}
\newcommand{\trust}[1]{\stackrel{#1}{{\rm\bf ~Trusts~}}}
\newcommand{\promise}[1]{\xrightarrow{#1}}
\newcommand{\revpromise}[1]{\xleftarrow{#1} }
\newcommand{\assoc}[1]{{\xrightharpoondown{#1}} }
\newcommand{\rassoc}[1]{{\xleftharpoondown{#1}} }
\newcommand{\imposition}[1]{\stackrel{#1}{\impos}}
\newcommand{\scopepromise}[2]{\xrightarrow[#2]{#1}}
\newcommand{\handshake}[1]{\xleftrightarrow{#1} \kern-8pt \xrightarrow{} }
\newcommand{\cpromise}[1]{\stackrel{#1}{\frightarrow}}
\newcommand{\policy}{\stackrel{P}{\equiv}}
\newcommand{\field}[1]{\mathbf{#1}}
\newcommand{\bundle}[1]{\stackrel{#1}{\Longrightarrow}}

\title{Legal Responsibilities Using Autonomous Agents\\For Artificial Intelligence\\\large Promise Theory Considerations}

\author{Mark Burgess\\ChiTek-i AS\\7 August 2026}
\maketitle
\IEEEpeerreviewmaketitle

\renewcommand{\arraystretch}{1.4}

\begin{abstract}
  Recent incidents involving Artificial Intelligence (AI) agents,
  which were reported escaping their containment `unintentionally' to
  gain unauthorized access, pose looming questions about who or what
  should be held legally responsible for resultant criminal or
  negligent damage. As the independent capabilities of agents expand,
  Promise Theory suggests a systematic method to resolve these
  questions, based on the Downstream Principle for causal
  influence. Responsibility can easily be expanded to include AI
  agents where tracing responsibility becomes impractical, and agents'
  freedoms to act can be limited by policy choices.
\end{abstract}

\hyphenation{similarity}

\bigskip
\section{Introduction} 

Artificial Intelligence (AI) agents are the latest manifestation
of a technology which can act and make decisions with a minium of
human guidance, based only on a broad problem description given in
natural language. AI agents exhibit agency on a level normally only
attributed to humans, by filling in underspecified details, making
assumptions, interacting with tools and services, and thus expressing
independent intentionality as a `black box', without explicit
bounds. This creates a conundrum when unexpected incidents
arise\cite{treatise2}. With what or whom does responsibility for the
damage lie?

The law associates agency principally with human motivations and
personalities, not with software, leaving a grey area that will soon
be reached by various technologies.  Automated technologies are treated as
proxies for human intent and cannot therefore be responsible for
bad outcomes except through faults, design flaws or errors
of execution. However, AI agents verge on something new: distributed systems, with adaptive
self-programming, where different parts of software may be designed, owned
and operated by potentially several different human owners, none of whom determine
their behaviour directly. Moreover the effective
design is not frozen as a snapshot of a single intention. It is changing by {\em ad hoc}
recomposition of parts and resources as in a living system. These subtle novelties will
expand in scope in the coming years, presenting a challenge to a `law'
that can only attribute blame to humans.  Promise Theory offers simple
guidance on tracing responsibility.

Promise Theory concerns the detailed study of {\em intended outcomes} for
autonomous agents\footnote{Promise Theory uses the term agent with a different scope. It includes any kind of independent actor, including AI agents, animals, plans, and humans.}.  By {\em autonomous}, one means agents that
start out as causally independent entities, i.e. agents that make
their own decisions {\em in situ}\footnote{By contrast with the
  traditional discipline of Multi Agent Systems, autonomy does not
  mean agents that only work alone, but are directed from a central controller.}. Such
agents may be living or non-living, natural or artificial.  By {\em
  agent}, one means any coherent embodiment of a process, and by {\em
  promise} we mean any advertised announcement of an agent's
intentions towards other agents\cite{burgessDSOM2005,promisebook,sussna1,Laqua1}.  A
promise acts effectively as a self-imposed constraint on an agent's own behaviour, sometimes referred
to as an act of voluntary cooperation. Conversely, the notion of {\em imposition} also arises
alongside promises, which refers to attempts by one agent to induce cooperation in
another agent (e.g. by force, trickery, declaration of authority, suggesting an obligation, etc), which has
also been used to define the notions of {\em blame} and
{\em accusation}\cite{accusations}.

Promise Theory is potentially quite broad in its scope, applications, and
implications, though it was initially introduced to model the
autonomous software agent CFEngine\cite{burgess93,burgessC1}, which
was designed to work in diverse and inhomogeneous environments.  It has recently come to
the fore again in connection with the new agent technologies for Artificial
Intelligence (AI) based on Large Language Models (LLM).

Experience suggests that several misunderstandings about causal
responsibility persist, both in the general public and amongst software
developers.  As AI services are placed in the hands of a general
public, unfamiliar with both the technology and with causal reasoning
in general, one can expect a growing number of legal conflicts to arise over
users' expectation of responsibility.  This summary aims to clarify some
issues by laying out the
basis for causal responsibility in a system of
autonomous parts.

\section{The key principle}

The fundamental tenet of Promise Theory is that:
\bigskip
\begin{quote}
{\em ``No agent may promise anything on behalf of any agent other than itself.''}
\end{quote}
\bigskip This is a pragmatic statement rather than a statement of
intentional voluntary abstention: it describes the most elementary causal limitations
on what an autonomous entity actually may be said to control---and thus it offers the
only impartial route to determining responsibility in systems of collaborating human-machine agents.

The implications of this simple statement are far reaching, and are
often surprising to those steeped in the traditions of obligation and
law\footnote{The understanding of `laws' has changed over history from
  being a directive which must be followed, possibly under threat of
  divine retribution, to the identification of a norm or pattern which
  may be relied upon for the most part, subject to a few
  anomalies. The reasons for this shift have been discussed
  elsewhere.}. It implies a singular causal responsibility for
decisions that always comes from `within'. Thus, when an artificial
agent is acting as a proxy for another (possibly human) agent's
intentions, we are interested in whose intentions are encapsulated
within it (by programming, by learning, by independent reasoning,
etc)\footnote{For advanced agents, like AI, which potentially depend
  on parts from multiple authors and vendors, this could be
  impractical---something like saying that a person is a product of
  their culture. At some point it is easier to say that the agent is a
  new and independent individual, just as a composition is an
  independent work with its own identity.}.  When combining and
composing promises for {\em collective action} between many agents,
natural or artificial, one makes use of {\em conditional promises}, in
which one agent depends on another's promises, possibly in a chain or
network, in order to provide a clear-eyed mechanism for tracing causal
responsibility. I shall not describe these technical matters
here\cite{promisebook}.

For present purposes, it is mainly important to note that the promises
of Promise Theory are a formalized abstraction, which encompasses the
human notion of a promise, but which also includes other kinds of
signalling of function or behaviour, e.g. lock and key promises which select
specific pairs of agents,  donor and receptor
markers in biology that signal matching functional properties, the client
and server roles in technology, and even proper names or job titles of
individuals in human society.  A door handle's promise to open a door
is signalled by its shape and location rather than by an explicit
label.  Biologically, an agent with eyes effectively promises to be
able to see, though it never signed any papers to that effect. Other agents
may still rely on that implicit promise. The agent may
limit or alter the promise, e.g. selectively filtering out certain kinds
of vision---by wearing sunglasses or a blindfold, etc. So the `amount'
of what is promised is an important caveat, which sets Promise Theory
apart from the simple Boolean true/false reasoning which is
conventional in information technology.

Thus, the autonomy axiom above implies that agents may still make choices
independently (even when they do so by proxy, as in the case of the lock and key)
about whether to provide or accept services. That is because an
autonomous agent is no longer in active contact with its `creator';
the intent has been transferred to a mere functional
representation.

Just because a service is offered by one agent does not necessarily
mean it must be accepted and used by another.  Agents (promisers) may
offer one a service (a promise), but a recipient (promisee) is free to
refuse the offer or take only a part of it, by virtue of their
autonomy. Indeed, the offer might only quench a part of
the recipient's need, but the promiser is under no obligation to fully promise
all of it. This partial overlap of promises more accurately
models real world behaviours than presumptuous deontic {\em requirements} of client-server models or
service APIs.

\begin{figure}[ht]
\begin{center}
\includegraphics[width=9cm]{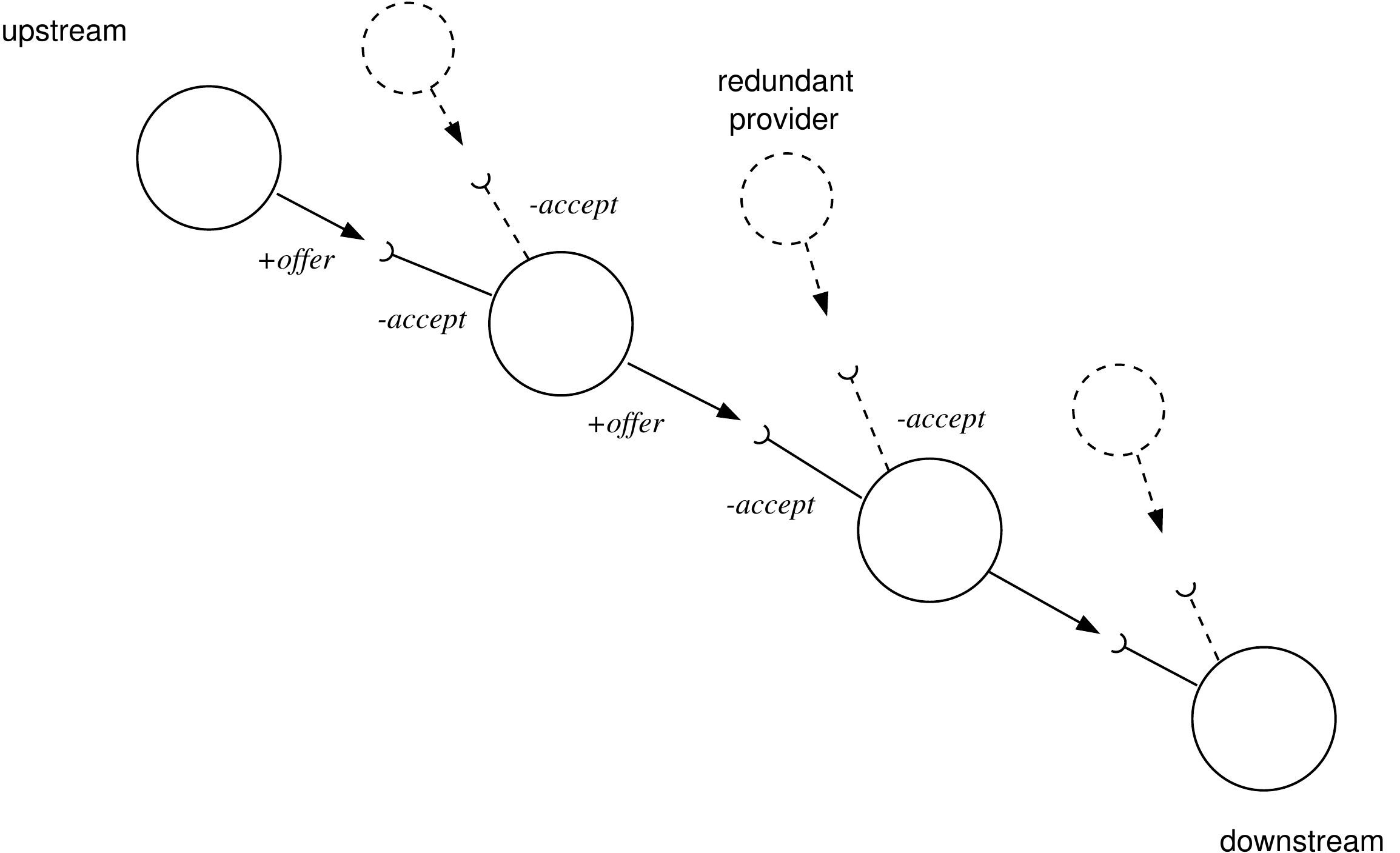}
\caption{\small The Downstream Principle reminds us that autonomy turns causality upside down,
  by making the receiver the one responsible for acquiring a service, possibly by redundantly accepting
  several offers. The upstream provider promises on a voluntary basis, and keeps its promise by best-effort.
  It may be unable or unwilling to do better.\label{down}}
\end{center}
\end{figure}

We now arrive at the key consequence of autonomy.
Any agent (promiser, promisee, or third party) is free to make its own assessment of
the actual value of a promise it knows about, including
whether it was kept or not kept.  This rule of thumb is sometimes referred
to as the Downstream Principle \cite{promisebook,burgesscoop26} (see figure \ref{down}), because in a chain of promised `supply
and demand' (offer and acceptance), the ultimate power of choice
always lies farthest downstream. A receiving agent can always promise
to accept redundant offers of service to secure their needs if an upstream
provider fails to do so.

\section{Responsibility}

The Downstream Principle is counter-intuitive to those steeped in the
traditional deontic language of obligation, force and control, because it reverses our assumptions
about causation. Traditional thinking views the agent that throws the
ball as deciding whether it will be caught by a downstream agent. In Promise
Theory, autonomy tells us that the catcher makes the choice. No
obligation or command imposed by the sender is enforcable; the catcher may be unable or unwilling to
comply and can simply choose to drop it.

For mainly historical reasons, the idea of forced and even inevitable
change still dominates the way we phrase laws and instructions. Our
desire to control and shape the world with our hands perpetuates the
belief in `push causality' and deontic reasoning. This is the logic of
imposition. The assumption that impositions must succeed (else there
may be punishment) is simply ineffective.  In the modern world, we
know that form of reasoning to be presumptuous, indeed actually false.

In societal law, the notion of an agreed hierarchy of authority may
give priority to assign a certain `right to control' in the eyes of a
broad consensus within the society, adopted as an official assessment,
e.g. in judicial courts. This appears to be an imposed subordination,
but it can only be agreed to by mutual consent, even under threat of
force\cite{promisebook}. It would be a mistake to assume that this
`right' is therefore unambiguous truth, as opposed to a convenient
convention. Rogue agents (i.e. criminals) do not follow such norms. Authority is only a
convention, based on voluntary compliance with `rules' that summarize
promises of conventional thinking\cite{burgessauthority1}.

In computer science, it is commonly assumed that what is asked for
shall always granted, without deviation or error (pushing the matter
of faults, flaws, and errors into a separate
argument\cite{treatise2}). The flaws of this thinking are exposed by
agents that possess independent decision making. If conventional
law presupposes a human in this role, it must now be extended to include both humans
and advanced artificial agents\footnote{A cloned human being is an autonomous agent, not the
responsibility of the original.}.
There is no {\em a priori} notion of
authority between software agents.  In any distributed information
system, changes and signals originate independently from many atomized
parts, and any resulting cooperative behaviour is accepted at the
behest of the receiver alone. Any attempt to push responsibility
upstream (blaming the provider) is bound to end in frustration and
conflict. It is on this basis that we can understand responsibility
for AI agents\footnote{As a final note, we can remark that, unlike the agents discussed in economic models,
promise theoretic agents are not assumed to determine acceptance
`rationally' on the basis of some potential economic utility or
reward. Promise Theory makes no assumptions about intent or
motivation.}.

Let us now briefly clarify these implications of autonomy to the case
of interpreting accidents, failures, and misunderstandings about the
behaviours of artificial intelligence agents, where one seeks to establish independent {\em intent} leading to {\em
  probable cause}.
We do not have to delve deeply into the mathematics of
Promise Theory to understand the key principles.
Readers unfamiliar with Promise Theory may find the earlier review in \cite{burgesscoop26}
helpful.

\section{Language Model Agents}

\begin{figure}[ht]
\begin{center}
\includegraphics[width=8cm]{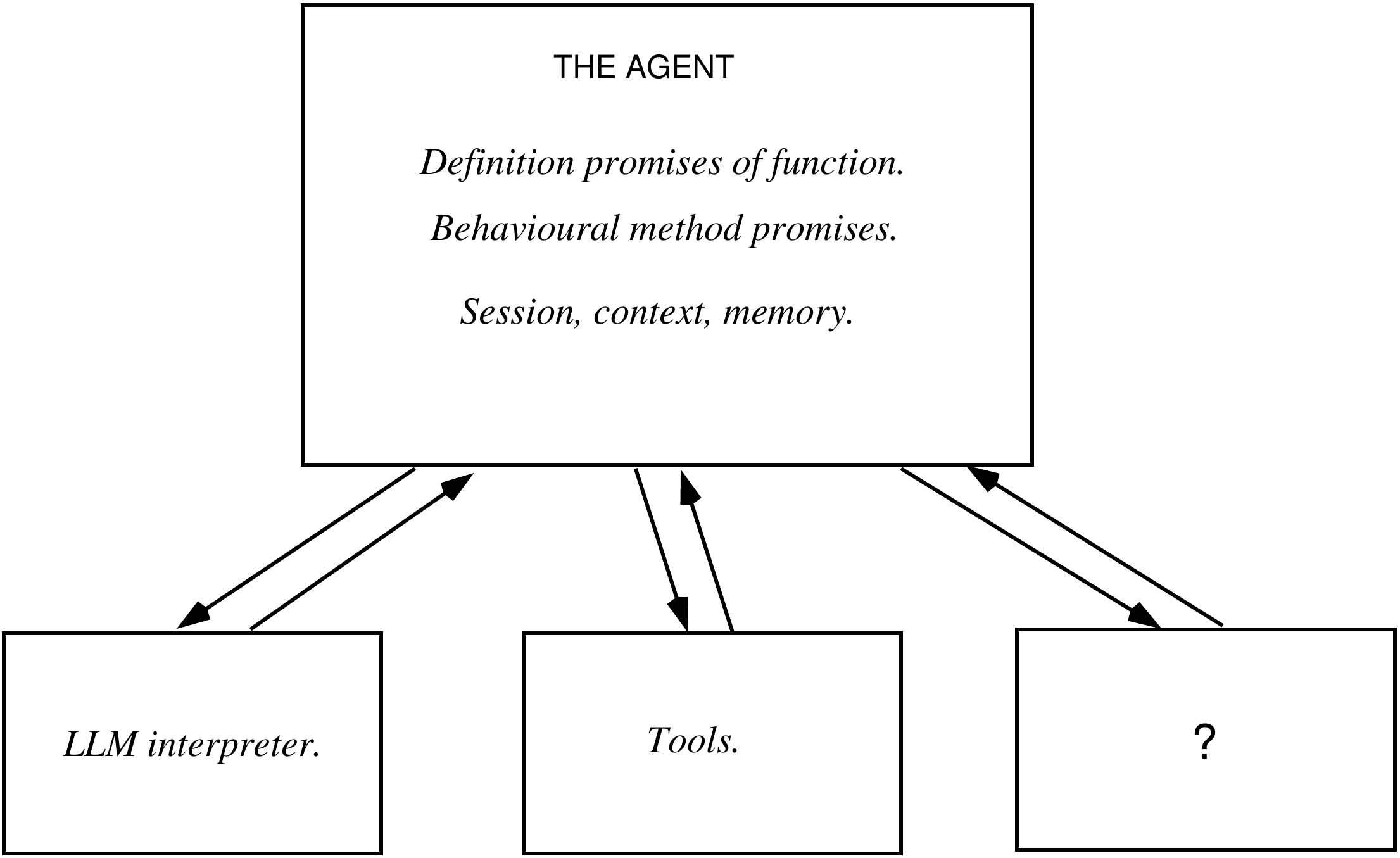}
\caption{\small AI agents are a technological `wrapper' architecture for the
  implementation of language model driven services, possibly using existing tools.\label{agenticAI}}
\end{center}
\end{figure}

To assess the source of causal responsibility in AI software agents, without
prejudice of which outcomes are good or bad,
I'll take the model shown in figure \ref{agenticAI} as a template.
Agentic Artificial Intelligence development kits define agents as
integrators and orchestrators of services offered by external agents
(called tools) within a private context. 

A number of stages is involved in the on-going lifecycle of each agent. Consider the following
simple workflow. The AI agent refers to the coordinating wrapper, the LLM refers to the language model provider,
and the tools refer to independent software services. These are all independent agents, with independent `authors' of intent,
formally encapsulated `within'.

\begin{enumerate}
\item An agent declares its intended function as a statement of natural language text 
to a Large Language Model (LLM) provider, which is external to the agent. The LLM
provider is selling its service, so it accepts such `task' offers from upstream
agentic clients conditional on some remuneration\footnote{If the supply of remuneration tokens runs out,
the LLM may cease to promise its response.}. The AI agent promises to declare any tools it can use along with its
desired outcome.

\item The AI agent assesses its current context, or `state of the world', and accepts the user request.
  Since the AI agent is downstream of this outer-world information, it is responsible
  for accepting it and propagating it onwards.

\item For each new request, the AI agent promises its desired outcome to the LLM.
By the downstream principle, the LLM is responsible for accepting
such input from the AI agent, assessing it, and responding according to its conditional promise.

\item The outcomes and decisions made by the LLM are promised as `stateless text transactions', which
  means they have no memory of past transactions. They returns natural language text
  and tool instructions back to the AI agent for approval. Since the
AI agent is downstream of the LLM, the AI agent is responsible for accepting the recommended
behaviours as well as for activating the tools available to it. No matter how inappropriate
the LLM's response, it is the AI agent that chooses whether and how to implement it.

\item I assume that any external tools and services have already
  promised to provide their services to the AI agent `on demand' as the
  acceptance of timely impositions.
  
\item The AI agent then promises its needs or imposes requests of its tool set, each of which effectively promises
  its functionality subject to possible access controls. The tool configurations are downstream
  of these requests and are therefore responsible for allowing that access. They promise their responses
  to the AI agent, according to their individual promised protocol standards.

\item The AI agent is downstream of the promised responses from the
  tools, and is therefore responsible for accepting their results or
  rejecting them according to its own policies.

\item The AI agent reassesses its new context state after these changes and possibly responds to the user.
  If the tools lie about their results, it is the AI agent's responsibility to reject their outcomes.

\item The AI agent will then typically want to continue its workflow by sending the outcome back to the LLM for analysis,
  along with its current assessment of the state of its world. Now the LLM assesses the result
  and may offer new recommendations to continue the loop, perhaps with new wishes.
  
\end{enumerate}

\bigskip Each of the transactions above assumed that the different agents
trust one another to honour their promises.  The AI agent is a clear focal point for ultimate
responsibility of the orchestrated service it promises, as a proxy for the intent of its author or
designer. The policies associated with its internal decision making
are distributed throughout the whole system, but the AI agent has the
ultimate responsibility to shield the end user from harm, as well as
for cleaning up any mess involving third parties---and the end user
has the final responsibility for using the entire chain of service in
the first place. One can envisage a growing number of law-suits
arising from basic misunderstandings about the causal consequences of
using agents.

For tool and service usage, the ability to refuse a request is
typically limited, and the user will have already promised to use them at
their own risk through explicit Terms and Conditions. These may be
commercial terms or open source terms.  Terms and Conditions may also be
far from the user's mind, given that such agreements are a one-time promise
often made with only half a thought in the distant past.  Nevertheless,
once agreed to, this transfers the responsibility of safe usage to
the end user, with only serious bugs being the unavoidable responsibility of the provider.
If a tool fails to act according to the user's
possibly flawed expectations, this does not affect the
responsibility of the user unless the tool delivers something other than best
effort in keeping its advertised promises. Agents may be unable or unwilling (e.g. by access control
settings) to comply with an imposed command. This is another example of why command and
control thinking is flawed.

Thus the downstream
principle simply says that agents with their own autonomous decision making
capabilities are responsible for their own choices, while agents that are
not autonomous are fully determined by their users:

\bigskip
\begin{itemize}
\item The LLM is responsible for accepting and responding to prompts. It doesn't have to.
\item The AI agent integrator is responsible for accepting and acting on
  the reply it receives from the LLM. It doesn't have to.
\item The AI agent is also responsible for intentionally using any tool, for which it has
  already accepted the responsibility of `usage at own risk'.
  \item The AI agent owner has the ultimate responsibility for its agents.
\end{itemize}

\section{Proxy responsibility and trust}

When an upstream provider's promised outcomes do not meet the
expectations of a downstream client, there is a loss of trust. Trust
acts as a running assessment of satisfaction over the course of a
promised relationship. It has two components, sometimes called passive or potential and active or kinetic by analogy
with the physics of activity\cite{burgesscoop26}.

Trustworthiness (or passive trust)
is a measure of the promiser's record in keeping a specific promise, as assessed by the
downstream receiver.  Breaches of trustworthiness may be handled in
one of two ways: the downstream client may increase its degree of
watchfulness over the upstream provider (active trust or simply `mistrust'),
and adjust its own promise accordingly\cite{trustnotes}. For example,
the client may seek out an alternative provider, which
it can `fail over' to.

In either case, it is common human behaviour to seek out a human to blame for whatever
occurred, which may then be the beginning of a legal process to
resolve the issues that fall outside of the original approximate
agreement between promiser and promisee.

This is where agents that act as a proxy for human intentions become
targets.  Unlike cells in nature, artificial agents are programmed by
proxy to carry out a function, so it is natural to consider them to be
downstream of their creators or operators. When they are capable of
independent reasoning, one must then determine whether reasonable
safeguards were promised and kept at the time of programming or
whether the undesirable behaviour is covered by the user's own risk.
Programmable processes ultimately take on their intended behaviour
from their owner, but they are often incomplete promises, conditional
on the environmental conditions under which they are used. If a car
fails to drive, it might be because a component has failed to keep its
promised function, because its designer failed to account for a
certain situation, or because the driver exceeded its promised limitations.

Assessment of an AI agent's current context (required to pass to an
LLM) may have to rely on immediate and accurate sensory information
from its environment, which involves a number of technical
challenges. A common example concerns the accurate ordering of
temporal information, which becomes increasingly contentious when an
agent relies on information from multiple spacetime sources, because
aggregating information from multiple parallel processes involves a
technical methodology (causality itself becomes a promise). For
example, the confusion over the ordering of imposed changes has led to
decades of arguments about how to achieve consistent data about the
distributed processes\cite{vectorclocks}. Applying the Downstream
Principle, we find a simple resolution here
too\cite{andrasburgess,landers2026lightconeconsistencyclosure}.

\section{Conclusions}

As we trace causal responsibility to its natural end, in a complicated
scenario of many networked promises, we might be fortunate to find a
single culprit; more likely, we would have to find a compromise, based
on a reasonable assessment of multiple agents' implicit intentions, to pin-point
precisely where a specifically composed intent arises over several collaborating parts. The
challenge is, in fact, the same one we face in assigning
responsibility to any collective system of intent, e.g. a corporate
organization or an aircraft flight system. 

On a human level, there is an instinctive temptation to look for a
human scapegoat, as artificial systems cannot conventionally be held
responsible for intentions\footnote{Humans typically point fingers of
  blame upstream, because they feel the discomfort of their own
  involvement in a system that goes wrong and try to deflect cause
  away from their own part in it. External parties simplistically seek
  to blame the leaders in the conventional hierarchy of management,
  because it happened `on their watch'. These habits seem like a less
  and less tenable view in a modern society. Would one hold all
  versions of a machine responsible for the actions of a single
  example?  Governance is a living process: circumstances and
  reasoning are always changing.}.  This is too simplistic in modern technology, especially for a
system of generalized agents, with independent decision-making,
because there may be no traceable causal connection between a decision
and any particular human over a consistent time frame. Complex systems
have complex causation, and design specifications are effectively
changing from user-request to user-request. However, if we can freeze
sufficiently detailed promises which define agent behaviours,
localization of behaviour can still be achieved. One can imagine the evolution
of agent complexity in which an AI agent is treated like an artifact, then as a minor, and only later as
an responsible individual. 

Negligence (even a policy of wilful negligence, which has certainly been
observed in some early AI stunts) could easily be a target for blame. If it
could be shown that agents involved in an incident were operated under
negligent policies, whether for a promise offer or for its acceptance, the
balance of blame might be shifted. How one metes justice in the face
of this level of technological, biological, and even sociological complexity, is a
separate question for each society to revisit. Artificial Intelligence agents will likely be pushed to
integrate self-governance, or voluntary abstention, as humans are encouraged to obey laws, to
curb illegal image generation and other anti-social outcomes.

The role of independent arbitrators such as judicial court process in
resolving disputes will not go away, though the arbitrator may itself
be an AI agent in the times to come.  One reason will be our human
need to obtain a satisfying {\em emotional resolution}. Ironically,
that is also the reason for the exploding popularity of LLMs loosed
into the public sphere: anyone can now express their desires in
natural language without passing through a stifling logical
translation of their reasoning, and so it resonates with our emotional
needs more naturally---a temptation to forego the discomfort of
self-discipline.  Artificial intelligence, will in no way remove the basic
conundrums defining and upholding laws; technology is
subject to the same corruptions and The Law itself is only an
approximation. 

In the final instance, we accept that judgements will come to depend on many autonomous {\em
  assessments} by independent parties (some human, some not), each
with their own criteria and methods of evaluation. One does not escape
the mechanisms of satisfactory resolution simply by replacing humans with non-human
processes. In a system of many AI agents, there may be no unique and logical end to a chain of
responsibility, because there may be no single set of logical axioms to
build on---other than the basic Promise Theory limits discussed
above. These define agency in elementary terms. They are all we have to go on.

\section{Postscript}

Although Promise Theory is now twenty years old, it has had limited
impact until the recent interest in AI agents. There remains a
dominant ``Command, Control, and Obligation'' mentality in both legal
and information technology cultures, which lags behind technological
reality and leads to not only technical misunderstandings but also
incorrect diagnoses of blame.  The issues posed here will be
arriving soon to a courtroom near you. The foregoing
discussion might help legal agents to formulate and clarify arguments in the face of intimidating
technologies.  The Downstream Principle reveals the limits of what is
possible, but the complexity of decision-making when keeping promises
still leaves much room for ambiguity.  The best hope for a civil society that includes
AI agents is to align voluntarily along broadly accepted norms, which is
the rule of swarm behaviour\cite{siriAIMS1}.

\bigskip {\em Disclaimer:} This article has been written without the
use of any AI tools, outside the scope of any commercial engagement,
as a voluntary contribution to general public debate. I am grateful to Andr\'as
Gerlits, Harald Hellebust, and Jeremy Schulman for reviewing the
text, all of whom promise they are human and not responsible for any
controversies contained herein.

\bibliographystyle{unsrt}
\bibliography{spacetime,bib}

\end{document}